\documentclass[conference]{IEEEtran}
\IEEEoverridecommandlockouts
\usepackage{cite}
\usepackage{amsmath,amssymb,amsfonts}
\usepackage{graphicx}
\usepackage{textcomp}
\usepackage{xcolor}
\usepackage{booktabs}
\usepackage{multirow}
\usepackage{algorithm}
\usepackage{algpseudocode}
\usepackage[font=small]{caption}
\def\BibTeX{{\rm B\kern-.05em{\sc i\kern-.025em b}\kern-.08em
    T\kern-.1667em\lower.7ex\hbox{E}\kern-.125emX}}
\begin{document}

\title{Few-Shot Ordinal Learning for Day-Wise Freshness Estimation with Hyperspectral Fish Images\\
}

\author{\IEEEauthorblockN{Kazi Nabiul Alam \textsuperscript{1}, Pooneh Bagheri-Zadeh \textsuperscript{1}, Akbar Sheikh-Akbari \textsuperscript{1}$^{\star}$}
\IEEEauthorblockA{\textsuperscript{1} \textit{School of Built Environment, Engineering and Computing,} \\
\textit{Leeds Beckett University, Leeds, United Kingdom.}\\
Correspondence: A.Sheikh-Akbari@leedsbeckett.ac.uk
}
}

\maketitle

\begin{abstract}
Non-destructive food quality assessment has increasingly benefited from hyperspectral imaging (HSI), which captures spectral signatures linked to biochemical changes during storage. Estimating day-wise freshness, however, remains challenging owing to strong inter-fillet variability and scarce labelled data per product. All existing deep learning approaches for HSI-based freshness prediction operate under full supervision, requiring densely annotated training sets that are costly to obtain at the individual-product level. We introduce, to the best of our knowledge, the first few-shot learning framework for HSI-based food quality estimation. Each fillet defines a distinct episodic task, and a CORAL-style ordinal prediction head captures the ranked nature of freshness progression through cumulative threshold modelling. Biologically grounded monotonicity and embedding smoothness constraints further guide predictions toward plausible trajectories. On a 16-day salmon HSI dataset under a strict unseen-fillet protocol, our method achieves a mean absolute error of 1.58 days and $\pm$2-day accuracy of 72.3\% with only three labelled days per fillet, substantially outperforming scalar regression and label-distribution baselines under an identical unseen-fillet protocol.
\end{abstract}


\section{Introduction}

Ensuring that perishable food products remain fresh is vital for consumer health, waste reduction, and efficient supply chain management. Conventional approaches to freshness evaluation, including sensory panels, chemical assays, and microbiological cultures, tend to be destructive, slow, and costly, restricting their scalability in industrial settings. Hyperspectral imaging (HSI)~\cite{Xiao2025DLFood, Yang2025HSIReview} offers a compelling alternative by merging spectroscopy with spatial imaging, thereby producing rich spectral--spatial signatures ~\cite{Shahrzad2025Fish} that reflect underlying biochemical changes during storage and spoilage.

Deep learning architectures, particularly convolutional neural networks (CNNs), have demonstrated considerable promise in extracting complex spectral--spatial patterns from HSI data for quality and safety inspection across a range of food products including fruits, vegetables, and meat~\cite{Yang2025HSIReview, Hong2024SpectralGPT}. Deep architectures incorporating spectral and spatial convolutions~\cite{Hong2024SpectralGPT} and ordinal regression heads~\cite{Polat2022OrdCNN} have shown promise for quality timeline prediction, while compact few-shot designs achieve competitive classification accuracy with substantially fewer parameters~\cite{Xi2022FSLClassCov}. A recurring limitation, however, is that these fully supervised models require large annotated datasets, something rarely available when freshness labels must be assigned per individual fillet at fine temporal granularity. Beyond data scarcity, most HSI-based studies frame freshness estimation~\cite{Yang2025HSIReview} as either standard regression or nominal classification, overlooking the inherent ordering among storage days.

Addressing data scarcity, few-shot learning reformulates training around task distributions sampled through episodic protocols, enabling generalization from very few labelled examples~\cite{Song2023FSLSurvey, Zeng2024FSLSurvey}. Well-known deep few-shot strategies, including prototypical, matching, relation, and gradient-based meta-learners, have proven effective under low-shot constraints~\cite{Li2022CrossDomFSL, Bai2022FSLAdaptive}. Within HSI, few-shot approaches have recently been applied to remote sensing and crop classification with limited labels~\cite{Xi2022FSLClassCov, Yang2022FSLAgri}.

Standard regression and classification objectives, however, fail to preserve ordinal relationships between storage days, potentially producing temporally inconsistent estimates. Ordinal regression addresses this by explicitly modelling ordered label spaces~\cite{Polat2022OrdCNN, Ord2Seq2023, Shin2023OrdinalDL}. Consistent Rank Logits (CORAL)~\cite{CORAL2020} decompose the ordinal problem into binary tasks with weight sharing, guaranteeing rank consistency, yet combining such methods with few-shot paradigms remains unexplored.

We propose a label-efficient episodic ordinal regression framework for day-wise freshness estimation from hyperspectral images. Rather than relying on support-conditioned metric comparison, it attains few-shot generalization through episodic task sampling combined with biologically grounded ordinal regularizers that keep predictions temporally consistent on fillets never seen during training. To the best of our knowledge, this is the first application of few-shot learning to HSI-based food quality or freshness prediction, where all prior deep learning solutions have operated under full supervision~\cite{Xiao2025DLFood, Yang2025HSIReview, Shahrzad2025Fish}. The episodic protocol is paired with CORAL-style ordinal heads and dual regularization: monotonicity on predicted days and embedding smoothness on learned representations. Experiments show this combination decisively outperforms scalar regression and distribution-based baselines under a strict unseen-fillet evaluation.

Our key contributions are:
\begin{itemize}
    \item The first episodic, label-efficient framework for HSI-based fish quality estimation, performing day-wise prediction on entirely unseen fillets.
    \item CORAL-style cumulative ordinal regression to explicitly encode the ranked structure of storage timelines.
    \item Applying biologically motivated dual regularization (monotonicity and embedding smoothness) to enforce temporally consistent freshness trajectories.
\end{itemize}

\section{Methodology}

\subsection{Problem Formulation and Episodic Framework}
Given hyperspectral cubes $x \in \mathbb{R}^{B \times H \times W}$ with freshness labels $y \in \{1, \dots, D\}$, we formulate day-wise estimation as few-shot ordinal regression. Each fillet defines a task $\mathcal{T}_i$ split into a $k$-day support set $\mathcal{S}_i$ and a query set $\mathcal{Q}_i$ of remaining days. Both contribute to training:
\begin{equation}
\min_{\theta} \; \mathbb{E}_{\mathcal{T}_i} 
\left[
\frac{1}{2}\left(
\mathcal{L}(\mathcal{S}_i; \theta) + \mathcal{L}(\mathcal{Q}_i; \theta)
\right)
+ \lambda_R \mathcal{R}(\mathcal{S}_i \cup \mathcal{Q}_i; \theta)
\right],
\label{eq:episodic}
\end{equation}
where $\mathcal{L}$ is the ordinal loss and $\mathcal{R}$ encompasses temporal regularization. Support and query samples pass through the \emph{same} shared network $f_\theta$ with identical weights, not separate branches; they differ only in role within the episodic objective, where support days fix the per-fillet temporal anchors over which the regularizers operate and query days provide the held-out supervision. Averaging over both sets prevents overfitting to the few support samples.

\begin{figure*}[htbp]
\centering
\includegraphics[width=0.99\linewidth]{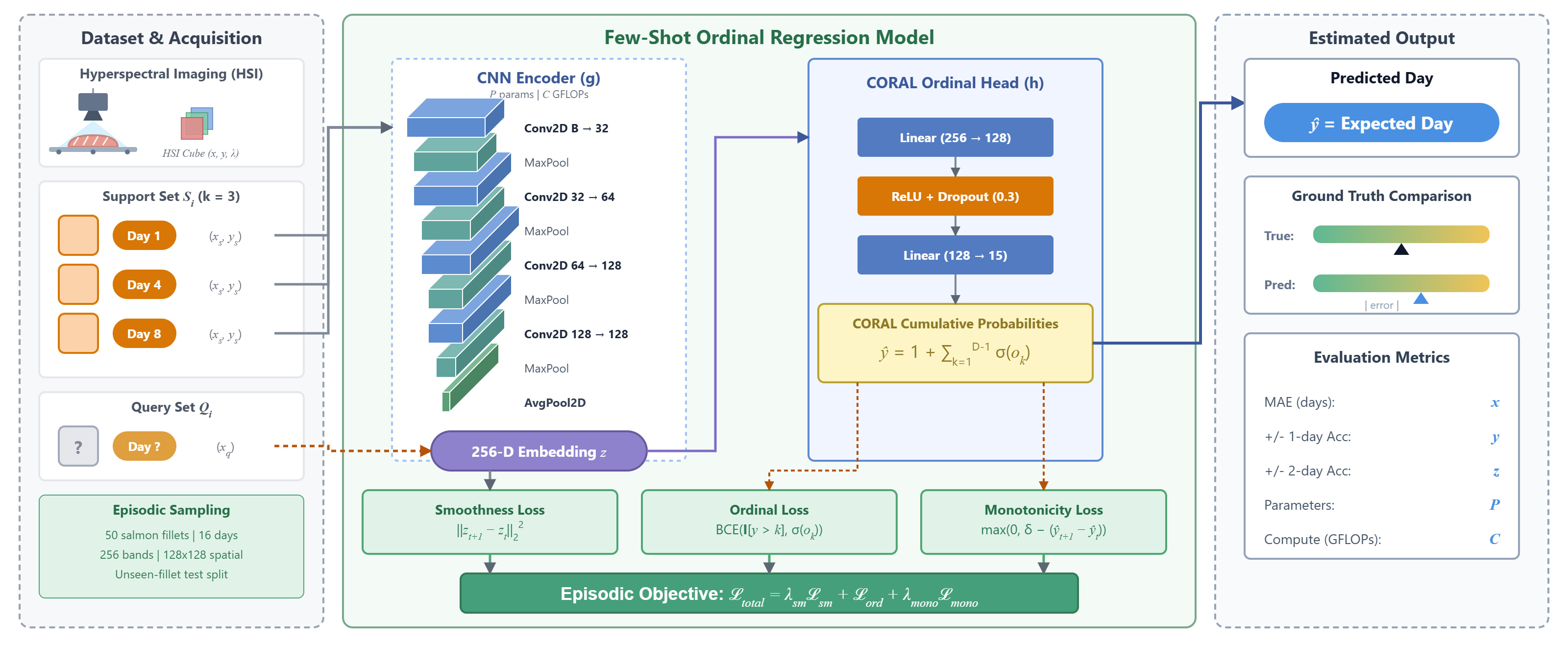}
\caption{Overall framework of the proposed episodic ordinal regression model. Support and query days share the \emph{same} encoder $g$ and ordinal head $h$ with identical weights, differing only in their role in the episodic objective (support days anchor the regularizers; query days provide held-out supervision).}
\label{fig:framework}
\end{figure*}

\subsection{Spectral-Channel CNN Backbone}
\label{sec:backbone}
Our backbone treats $B$ spectral bands as input channels to a 2D CNN rather than using 3D convolutions. Four convolutional blocks (32$\rightarrow$64$\rightarrow$128$\rightarrow$128 channels) with batch normalization, ReLU, and $2{\times}2$ max pooling extract hierarchical spatial features:
\begin{equation}
\mathbf{h}^{(l)} = \text{MaxPool}\big(\text{ReLU}(\text{BN}(\text{Conv2D}(\mathbf{h}^{(l-1)})))\big).
\end{equation}
Adaptive average pooling and a fully connected layer yield the embedding:
\begin{equation}
\mathbf{z} = \text{ReLU}(\mathbf{W}_{\text{fc}} \cdot \text{AdaptiveAvgPool}(\mathbf{h}^{(4)}) + \mathbf{b}_{\text{fc}}) \in \mathbb{R}^{d},
\end{equation}
with $d{=}256$. Cross-band correlations are learned implicitly through channel mixing. Table~\ref{tab:model_arch} presents the full architecture (441K parameters, 2.37 GFLOPs).

Treating spectral bands as input channels to a 2D CNN, rather than 3D or 1D spectral convolutions, is a deliberate trade-off: with $B{=}256$ bands and few labelled days per fillet, 3D convolution sharply increases parameters and overfits, whereas channel mixing in the first convolution still learns inter-band combinations jointly.

\begin{table}[htbp]
\centering
\caption{FreshnessOrdinal Network Architecture ($B{=}256$ Bands)}
\label{tab:model_arch}
\renewcommand{\arraystretch}{1.0}
\resizebox{\columnwidth}{!}{%
\begin{tabular}{lcc}
\toprule
Layer & Output Shape & Params \\
\midrule
\multicolumn{3}{l}{\textit{HSICNN Backbone}} \\
Conv2D ($B{\rightarrow}32$) + BN + Pool & $32 \times 64 \times 64$ & 133,088 \\
Conv2D ($32{\rightarrow}64$) + BN + Pool & $64 \times 32 \times 32$ & 18,624 \\
Conv2D ($64{\rightarrow}128$) + BN + Pool & $128 \times 16 \times 16$ & 74,112 \\
Conv2D ($128{\rightarrow}128$) + BN + Pool & $128 \times 8 \times 8$ & 147,840 \\
AdaptiveAvgPool + FC($128{\rightarrow}256$) & $256$ & 33,024 \\
\midrule
\multicolumn{3}{l}{\textit{CORAL Ordinal Head}} \\
FC($256{\rightarrow}128$) + ReLU + Drop(0.3) & $128$ & 32,896 \\
FC($128{\rightarrow}15$) & $15$ & 1,935 \\
\midrule
Total / GFLOPs / Memory & & 441K / 2.37 / 47MB \\
\bottomrule
\end{tabular}
}
\end{table}

% \begin{table}[htbp]
% \centering
% \caption{FreshnessOrdinal Network Architecture ($B{=}462$ Bands)}
% \label{tab:model_arch}
% \renewcommand{\arraystretch}{1.0}
% \begin{tabular}{lcc}
% \toprule
% Layer & Output Shape & Params \\
% \midrule
% \multicolumn{3}{l}{\textit{HSICNN Backbone}} \\
% Conv2D ($B{\rightarrow}32$) + BN + Pool & $32 \times 64 \times 64$ & 133,088 \\
% Conv2D ($32{\rightarrow}64$) + BN + Pool & $64 \times 32 \times 32$ & 18,624 \\
% Conv2D ($64{\rightarrow}128$) + BN + Pool & $128 \times 16 \times 16$ & 74,112 \\
% Conv2D ($128{\rightarrow}128$) + BN + Pool & $128 \times 8 \times 8$ & 147,840 \\
% AdaptiveAvgPool + FC($128{\rightarrow}256$) & $256$ & 33,024 \\
% \midrule
% \multicolumn{3}{l}{\textit{CORAL Ordinal Head}} \\
% FC($256{\rightarrow}128$) + ReLU + Drop(0.3) & $128$ & 32,896 \\
% FC($128{\rightarrow}15$) & $15$ & 1,935 \\
% \midrule
% Total / GFLOPs / Memory & & 441K / 2.37 / 47MB \\
% \bottomrule
% \end{tabular}
% \end{table}

\subsection{CORAL Ordinal Head and Loss}
\label{sec:coral}
For $D$ ordered classes, $D{-}1$ binary sub-tasks determine threshold exceedance. From $\mathbf{z}$, the head computes logits via a two-layer classifier with dropout:
\begin{equation}
\mathbf{o} = \mathbf{W}_2 \cdot \text{Dropout}\big(\text{ReLU}(\mathbf{W}_1 \mathbf{z} + \mathbf{b}_1)\big) + \mathbf{b}_2 \in \mathbb{R}^{D-1}.
\end{equation}
Cumulative exceedance probabilities $P(y > k \mid x) = \sigma(o_k)$ yield the expected day:
\begin{equation}
\hat{y} = 1 + \sum_{k=1}^{D-1} \sigma(o_k).
\end{equation}
Class probabilities are recoverable via $P(y{=}k \mid x) = P(y{>}k{-}1) - P(y{>}k)$ with boundary conditions $P(y{>}0)=1$, $P(y{>}D)=0$.

We adopt CORAL over a generic threshold-based model for two reasons: its shared-weight cumulative-logit construction guarantees rank-monotone thresholds by design, aligning with the strictly ordered storage-day labels and complementing the monotonicity regularizer of Eq.~(\ref{eq:mono}); and decomposing the $D$-way problem into $D{-}1$ binary sub-tasks keeps the head lightweight and well-conditioned under few labelled days, whereas unconstrained thresholds overfit and can violate rank consistency in low-data regimes.

Binary cross-entropy over all thresholds forms the ordinal loss:
\begin{equation}
\mathcal{L}_{\text{ord}} =
\frac{1}{N}\sum_{n=1}^{N}\sum_{k=1}^{D-1}
\text{BCE}\big(\mathbb{I}[y_n > k],\; \sigma(o_k^{(n)})\big),
\label{eq:ord}
\end{equation}
where targets $t_{n,k} = \mathbb{I}[y_n > k]$ encode the ordinal structure (e.g., $y{=}5$ gives $[1,1,1,1,0,\dots,0]$). Per-episode loss averages over support and query: $\mathcal{L}_{\text{episode}} = \frac{1}{2}(\mathcal{L}_{\text{ord}}^{\mathcal{S}} + \mathcal{L}_{\text{ord}}^{\mathcal{Q}})$.

\subsection{Temporal Regularization}
\label{sec:mono}
\textbf{Monotonicity.} For consecutive temporal pairs $\mathcal{P}$ sorted by ground-truth day across the combined episode:
\begin{equation}
\mathcal{L}_{\text{mono}} = \frac{1}{|\mathcal{P}|}\sum_{(t,t+1) \in \mathcal{P}}
\max\big(0,\; \delta - (\hat{y}_{t+1} - \hat{y}_t)\big),
\label{eq:mono}
\end{equation}
with margin $\delta{=}0.01$, enforcing biologically plausible ascending predictions.

\label{sec:smooth}
\textbf{Embedding smoothness.} Abrupt representational jumps between adjacent days are penalized on the $d$-dimensional embeddings:
\begin{equation}
\mathcal{L}_{\text{smooth}} = \frac{1}{|\mathcal{P}|}\sum_{(t,t+1) \in \mathcal{P}}
\left\| \mathbf{z}_{t+1} - \mathbf{z}_t \right\|_2^2.
\end{equation}
While $\mathcal{L}_{\text{mono}}$ constrains the output space, $\mathcal{L}_{\text{smooth}}$ shapes the representation space, forming a complementary pairing.

\subsection{Training Objective and Inference}
The per-episode loss combines all terms:
\begin{equation}
\mathcal{L}_{\text{total}} =
\mathcal{L}_{\text{episode}}
+ \lambda_{\text{mono}} \mathcal{L}_{\text{mono}}
+ \lambda_{\text{smooth}} \mathcal{L}_{\text{smooth}},
\label{eq:total}
\end{equation}
with $\lambda_{\text{mono}}{=}\lambda_{\text{smooth}}{=}0.1$. Adam optimization ($\text{lr}{=}3{\times}10^{-4}$, weight decay $5{\times}10^{-4}$) runs for 40 epochs of 60 episodes. Network parameters are initialized randomly using the standard He scheme for convolutional and linear layers, with no transfer learning or auxiliary pre-training, so results reflect training from scratch on the target HSI data alone. 

Fig.~\ref{fig:framework} and Algorithm~\ref{alg:training} describe the overall architecture of the proposed framework.

\begin{algorithm}[t]
\caption{Few-Shot Ordinal Meta-Training}
\label{alg:training}
\begin{algorithmic}[1]
\Require Train fillets $\mathcal{F}_{\text{train}}$, val episodes $\mathcal{E}_{\text{val}}$, support size $k$
\Ensure Trained parameters $\theta^*$
\State Initialize $\theta$; $\text{best\_mae} \leftarrow \infty$
\For{epoch $= 1$ to $E$}
    \For{episode $= 1$ to $M$}
        \State Sample fillet $f \sim \mathcal{F}_{\text{train}}$ (with $\geq N_{\min}$ days)
        \State Split into $\mathcal{S}$ ($k$ days) and $\mathcal{Q}$ (remaining)
        \State Compute $\mathcal{L}_{\text{total}}$ via Eqs.~(\ref{eq:ord})--(\ref{eq:total})
        \State Update $\theta$ via Adam
    \EndFor
    \State Evaluate on $\mathcal{E}_{\text{val}}$; save $\theta$ if MAE improves
\EndFor
\State \Return $\theta^*$
\end{algorithmic}
\end{algorithm}

\section{Results and Discussion}
\label{sec:results}

\subsection{Dataset and Experimental Protocol}
\label{sec:dataset}
\begin{figure*}[htbp]
\centering
\includegraphics[width=0.90\textwidth]{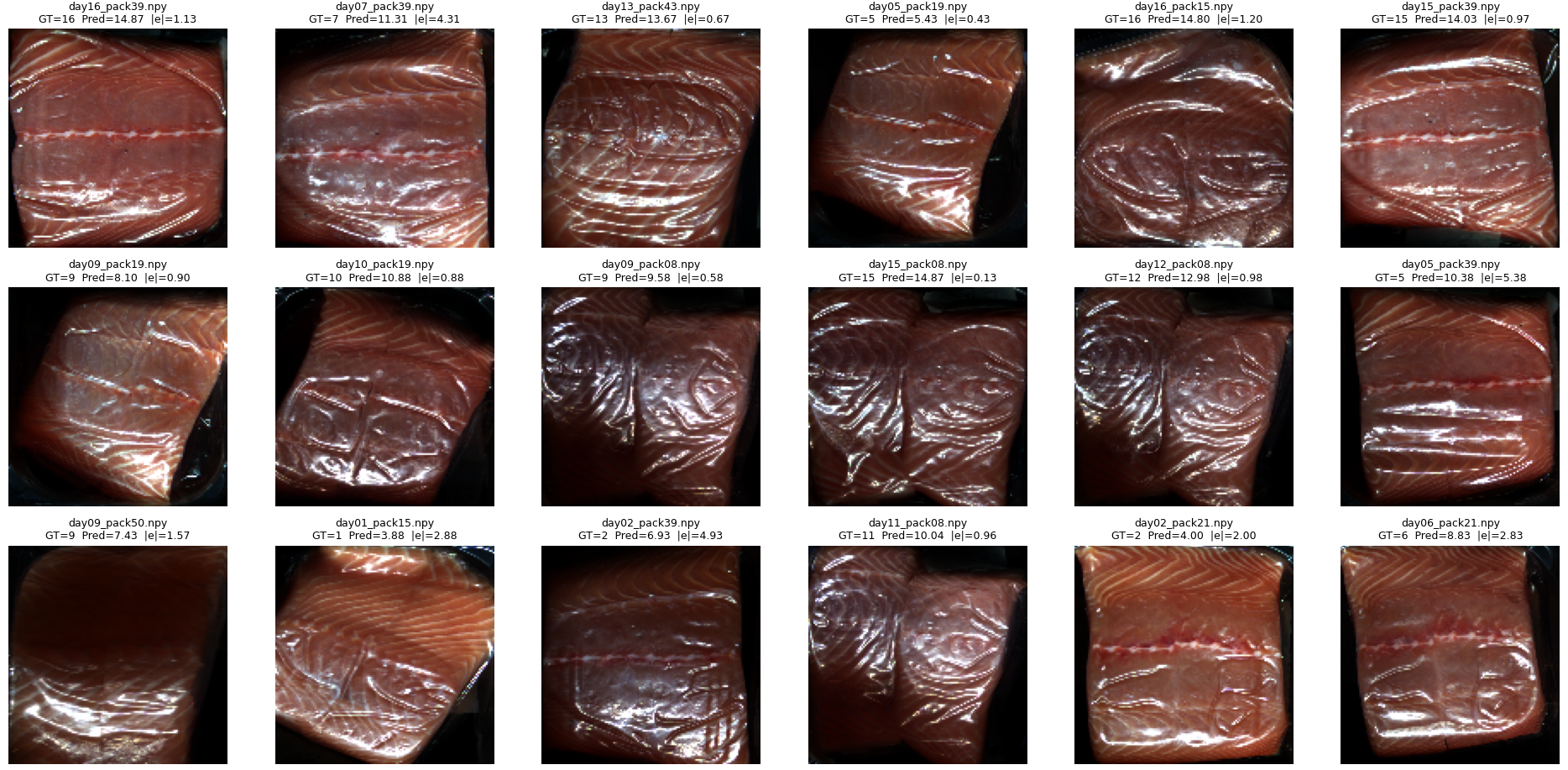}
\caption{Sample-level predictions on unseen test fillets with HSI$\rightarrow$RGB composites. Each panel shows ground-truth day (GT), predicted day (Pred), and absolute error ($|e|$); most fall within $\pm$2 days.}
\label{fig:check}
\end{figure*}

We evaluate on our custom-made salmon freshness HSI dataset of 50 individually packaged fillets imaged daily over 16 days ($D{=}16$; day~6 = labelled expiry). Each cube spans 462 spectral bands (386.88--1003.6~nm) at $128{\times}128$ resolution; after band-wise z-score normalization and noisy edge-band removal, $B{=}256$ channels are retained. HSI-to-RGB composites (Fig.~\ref{fig:check}) confirm that predictions typically fall within $\pm$2 days.

Splitting is performed at the \emph{pack level} to prevent data leakage (Table~\ref{tab:data_split}). The 30 training fillets ensure broad inter-fillet variability for robust episodic meta-learning, while 10 validation and 10 test fillets give low-variance metric estimates without leaking pack-specific cues.

Validation and test episodes use fixed seeds for reproducibility, considering only fillets with ${\geq}6$ available days. Each episode randomly selects $k{=}3$ support days and assigns the remainder as queries. Performance is reported as MAE, $\pm$1-day, and $\pm$2-day accuracy. Table~\ref{tab:hyperparams} lists all hyperparameters.

\begin{table}[!t]
\centering
\caption{Pack-Level Data Split (No Leakage Across Partitions)}
\label{tab:data_split}
\renewcommand{\arraystretch}{1.0}
\begin{tabular}{lccc}
\toprule
Partition & Pack IDs & Packs & Cubes \\
\midrule
Training & 1--30 & 30 & 480 \\
Validation & 31--40 & 10 & 160 \\
Test & 41--50 & 10 & 160 \\
\midrule
Total & 1--50 & 50 & 800 \\
\bottomrule
\end{tabular}
\end{table}

\vspace{-10pt}

\begin{table}[t]
\centering
\caption{Training Hyperparameters}
\label{tab:hyperparams}
\renewcommand{\arraystretch}{1.0}
\begin{tabular}{lclc}
\toprule
Parameter & Value & Parameter & Value \\
\midrule
Optimizer & Adam & $\lambda_{\text{mono}}$ & 0.1 \\
Learning rate & $3{\times}10^{-4}$ & $\lambda_{\text{smooth}}$ & 0.1 \\
Weight decay & $5{\times}10^{-4}$ & Margin $\delta$ & 0.01 \\
Epochs & 40 & Embedding $d$ & 256 \\
Episodes/epoch & 60 & Classes $D$ & 16 \\
Support $k$ & 3 & Dropout & 0.3 \\
Min. days $N_{\min}$ & 6 & & \\
\bottomrule
\end{tabular}
\end{table}

\subsection{Comparison with Regression and Distribution Methods}
Table~\ref{tab:baseline_comparison} benchmarks our approach against regression and distribution-based formulations on unseen test fillets. Scalar regression struggles with inter-fillet variability. Label distribution learning~\cite{Wen2023LDL} captures day-label uncertainty but lacks explicit ordinal encoding \cite{Castagnos2022SoftOrd}. Our method reduces MAE by 19\% over few-shot regression and boosts $\pm$2-day accuracy by 15.4 percentage points.

\begin{table}[t]
\centering
\caption{Comparison with Regression and Distribution-Based Methods with our approach}
\label{tab:baseline_comparison}
\renewcommand{\arraystretch}{1.0}
\begin{tabular}{lccc}
\toprule
Method & MAE $\downarrow$ & $\pm$1 Acc. $\uparrow$ & $\pm$2 Acc. $\uparrow$ \\
\midrule
\multicolumn{4}{l}{\textit{Regression Baselines}} \\
Linear Reg. (HSI feat.) & 2.87 & 21.4\% & 39.2\% \\
CNN + L1 Regression & 2.21 & 30.8\% & 52.3\% \\
Few-Shot CNN Reg. & 1.95 & 34.6\% & 56.9\% \\
\midrule
\multicolumn{4}{l}{\textit{Soft-Label / Distribution Baselines}} \\
Gaussian Label Smooth. & 2.04 & 31.5\% & 58.5\% \\
Label Dist. Learning & 1.86 & 33.1\% & 62.3\% \\
LDL + Temporal Smooth. & 1.79 & 35.4\% & 64.6\% \\
\midrule
\textbf{Proposed} & \textbf{1.58} & \textbf{42.3\%} & \textbf{72.3\%} \\
\bottomrule
\end{tabular}
\end{table}

\begin{table}[!t]
\centering
\caption{Impact of Ordinal Modelling over multiple heads}
\label{tab:ordinal_effect}
\renewcommand{\arraystretch}{1.0}
\begin{tabular}{lccc}
\toprule
Head & MAE $\downarrow$ & $\pm$1 Acc. $\uparrow$ & $\pm$2 Acc. $\uparrow$ \\
\midrule
Scalar Regression & 1.95 & 34.6\% & 56.9\% \\
Ordinal (no reg.) & 1.73 & 38.5\% & 66.2\% \\
\textbf{Ordinal + Reg.} & \textbf{1.58} & \textbf{42.3\%} & \textbf{72.3\%} \\
\bottomrule
\end{tabular}
\end{table}

\subsection{Effect of Ordinal Modelling and Support Size}
Table~\ref{tab:ordinal_effect} isolates the benefit of ordinal learning. Switching from scalar to ordinal prediction drops MAE from 1.95 to 1.73; adding regularization further reduces it to 1.58. Table~\ref{tab:fewshot_k} highlights robustness across support sizes ($k$): $k{=}3$, and even 1-shot achieves 48.5\% $\pm$2-day accuracy.

\begin{table}[htbp]
\centering
\caption{Robustness Across Few-Shot Support Sizes}
\label{tab:fewshot_k}
\renewcommand{\arraystretch}{1.0}
\begin{tabular}{cccc}
\toprule
$k$ & MAE $\downarrow$ & $\pm$1 Acc. $\uparrow$ & $\pm$2 Acc. $\uparrow$ \\
\midrule
1 & 2.34 & 26.2\% & 48.5\% \\
2 & 1.92 & 34.1\% & 61.5\% \\
3 (ours) & \textbf{1.58} & \textbf{42.3\%} & \textbf{72.3\%} \\
5 & 1.63 & 42.6\% & 72.1\% \\
\bottomrule
\end{tabular}
\end{table}
\vspace{-8pt}

% \begin{table}[!t]
% \centering
% \caption{Ablation Study on multiple configurations}
% \label{tab:ablation}
% \renewcommand{\arraystretch}{1.0}
% \begin{tabular}{lccc}
% \toprule
% Configuration & MAE $\downarrow$ & $\pm$1 Acc. $\uparrow$ & $\pm$2 Acc. $\uparrow$ \\
% \midrule
% A2: Ordinal Only & 2.29 & 29.2\% & 56.6\% \\
% A3: + Emb. Smooth. & 2.22 & 28.9\% & 56.8\% \\
% A4: + Monotonicity & \textbf{2.01} & \textbf{31.1\%} & \textbf{59.2\%} \\
% A5: + Both & 2.08 & 29.1\% & 57.2\% \\
% \bottomrule
% \end{tabular}
% \end{table}

\begin{figure*}[htbp]
\centering
\includegraphics[width=\linewidth]{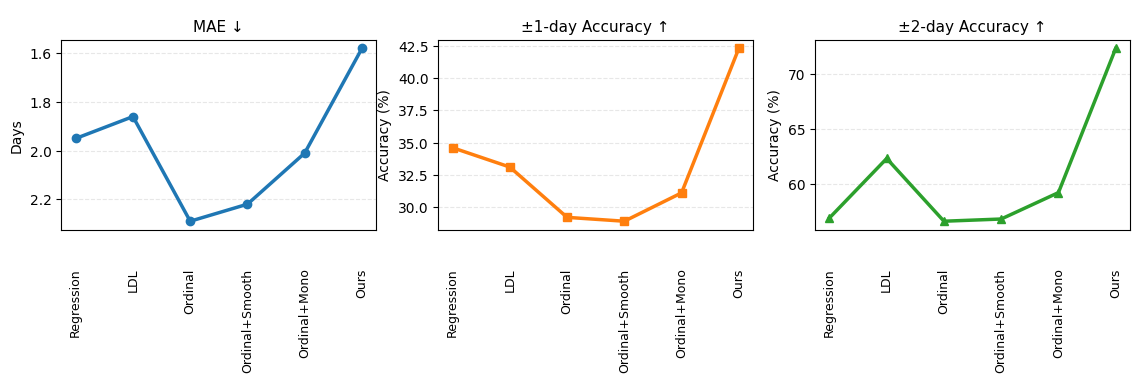}
\caption{Comparison of freshness estimation methods across MAE and tolerance-based accuracy metrics on unseen test fillets.}
\label{fig:comparison}
\end{figure*}

\subsection{Ablation Study}
Table~\ref{tab:ablation} reports component contributions under a reduced 15-epoch budget, adopted to hold optimization cost fixed across A2--A5 so per-component differences are comparable without convergence confounds.

\begin{table}[htbp]
\centering
\caption{Ablation Study on multiple configurations}
\label{tab:ablation}
\renewcommand{\arraystretch}{1.0}
\begin{tabular}{lccc}
\toprule
Configuration & MAE $\downarrow$ & $\pm$1 Acc. $\uparrow$ & $\pm$2 Acc. $\uparrow$ \\
\midrule
A2: Ordinal Only & 2.29 & 29.2\% & 56.6\% \\
A3: + Emb. Smooth. & 2.22 & 28.9\% & 56.8\% \\
A4: + Monotonicity & \textbf{2.01} & \textbf{31.1\%} & \textbf{59.2\%} \\
A5: + Both & 2.08 & 29.1\% & 57.2\% \\
\bottomrule
\end{tabular}
\end{table}

Absolute MAE is therefore higher than in Tables~\ref{tab:baseline_comparison} and~\ref{tab:ordinal_effect} and only relative ordering matters here. Monotonicity (A4) yields the largest single gain (about 12\% over the ordinal-only A2), matching the monotone progression of degradation, while smoothness alone (A3) helps only marginally. Under this short budget, A5 (2.08) does not improve over A4 (2.01), which we attribute to smoothness needing more episodes to stabilize the representation space; this is consistent with the full-budget Table~\ref{tab:ordinal_effect}, where the complete regularized model attains the best MAE (1.58). The ablation should thus be read as a relative-ranking diagnostic, not the final operating point, visually mentioned in Fig.\ref{fig:comparison}.

\vspace{-5pt}

\subsection{Qualitative Analysis}
\begin{figure}[htbp]
\centering
\includegraphics[width=0.98\linewidth]{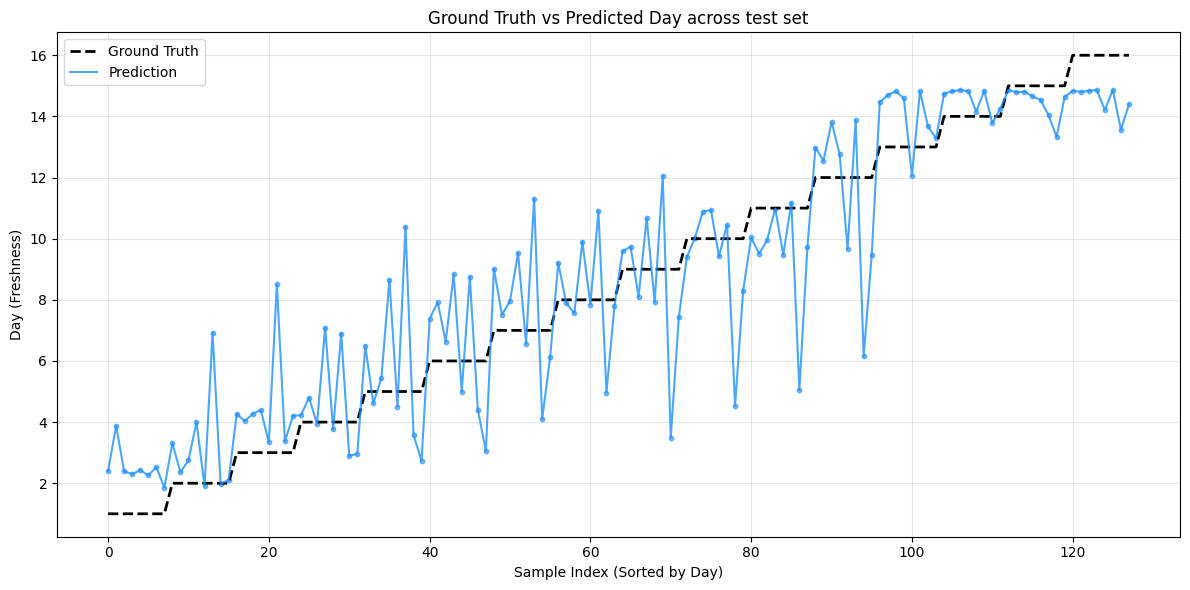}
\caption{Ground truth versus predicted freshness day across the test set. Predictions closely follow the ascending trajectory.}
\label{fig:gt_pred}
\end{figure}

Fig.~\ref{fig:gt_pred} illustrates our method's superiority across all metrics, with the sharp upward trend in $\pm$2-day accuracy confirming the value of dual regularization. Mapping predictions against ground-truth days shows tight adherence to the ascending trajectory, with larger errors confined to mid-range days (5--9) where biochemical changes are subtlest.
\vspace{-5pt}
\vspace{-5pt}
\section{Conclusion}

This paper proposes a few-shot ordinal learning paradigm for estimating day-wise freshness from hyperspectral images. The approach combines episodic training, CORAL-based ordinal regression, and temporal regularization to predict freshness across previously unseen fish fillets with only three labelled storage days per task. Experiments demonstrate consistent and significant improvements over label distribution learning and conventional regression under a rigorous unseen-fillet evaluation. A limitation is that the salmon HSI dataset is proprietary and not yet publicly available, which constrains reproducibility; the framework itself is dataset-agnostic, and we plan to validate it on public HSI food-quality benchmarks and release the code in future work. The method offers a label-efficient and configurable approach for scalable, non-destructive food quality assessment for industry.


\begin{thebibliography}{99}
\setlength{\itemsep}{0pt}
\setlength{\parskip}{0pt}
\footnotesize

\bibitem{Xiao2025DLFood}
Y.~Xiao \emph{et al.},
``Deep learning--based regression of food quality attributes using near-infrared spectroscopy and hyperspectral imaging: A review,''
\emph{Food Chem.}, vol.~493, pp.~145932, 2025.


\bibitem{Yang2025HSIReview}
C.~Yang \emph{et al.},
``Hyperspectral imaging and deep learning for quality and safety inspection of fruits and vegetables: A review,''
\emph{J. Agric. Food Chem.}, vol.~73, no.~17, pp.~10019--10035, 2025.

\bibitem{Shahrzad2025Fish}
F.~Shahrzad, Z.~Arabi, S.~Ghafari, and A.~Sheikh-Akbari,
``Fish quality assessment using hyperspectral imaging and computer vision: A review,''
\emph{IEEE Sensors J.}, vol.~25, no.~14, pp.~26255--26268, 2025.

\bibitem{Hong2024SpectralGPT}
D.~Hong \emph{et al.},
``SpectralGPT: Spectral remote sensing foundation model,''
\emph{IEEE Trans. Pattern Anal. Mach. Intell.}, vol.~46, no.~8, pp.~5227--5244, 2024.

\bibitem{Xi2022FSLClassCov}
B.~Xi \emph{et al.},
``Few-shot learning with class-covariance metric for hyperspectral image classification,''
\emph{IEEE Trans. Image Process.}, vol.~31, pp.~5079--5092, 2022.

\bibitem{Song2023FSLSurvey}
Y.~Song, T.~Wang, P.~Cai, S.~K. Mondal, and J.~P. Sahoo,
``A comprehensive survey of few-shot learning: Evolution, applications, challenges, and opportunities,''
\emph{ACM Comput. Surv.}, vol.~55, no.~13s, pp.~1--40, 2023.

\bibitem{Zeng2024FSLSurvey}
W.~Zeng and Z.-Y. Xiao,
``Few-shot learning based on deep learning: A survey,''
\emph{Math. Biosci. Eng.}, vol.~21, no.~1, pp.~679--711, 2024.

\bibitem{Li2022CrossDomFSL}
Z.~Li \emph{et al.},
``Deep cross-domain few-shot learning for hyperspectral image classification,''
\emph{IEEE Trans. Geosci. Remote Sens.}, vol.~60, pp.~5501618, 2022.

\bibitem{Bai2022FSLAdaptive}
J.~Bai \emph{et al.},
``Few-shot hyperspectral image classification based on adaptive subspaces and feature transformation,''
\emph{IEEE Trans. Geosci. Remote Sens.}, vol.~60, pp.~5518517, 2022.

\bibitem{Yang2022FSLAgri}
J.~Yang \emph{et al.},
``A survey of few-shot learning in smart agriculture: Developments, applications, and challenges,''
\emph{Plant Methods}, vol.~18, no.~1, pp.~28, 2022.

\bibitem{Polat2022OrdCNN}
G.~Polat, I.~Ergenc, H.~T. Ozlu, and A.~A. Alatan,
``Class-distribution-aware calibration for long-tailed visual recognition via ordinal regression,''
in \emph{Proc. IEEE/CVF CVPR Workshops}, 2022, pp.~2873--2882.

\bibitem{Ord2Seq2023}
J.~Wang \emph{et al.},
``Ord2Seq: Regarding ordinal regression as label sequence prediction,''
in \emph{Proc. IEEE/CVF ICCV}, 2023, pp.~11161--11171.

\bibitem{CORAL2020}
W.~Cao, V.~Mirjalili, and S.~Raschka,
``Rank consistent ordinal regression for neural networks with application to age estimation,''
\emph{Pattern Recognit. Lett.}, vol.~140, pp.~325--331, 2020.

\bibitem{Shin2023OrdinalDL}
V.~Vargas, P.~A. Guti\'{e}rrez, and C.~Herv\'{a}s-Mart\'{i}nez,
``Unimodal regularisation based on beta distribution for deep ordinal regression,''
\emph{Pattern Recognit.}, vol.~122, pp.~108310, 2022.

\bibitem{Wen2023LDL}
T.~Wen, B.~Yang, J.~Wang, and G.~Chen,
``Label distribution learning by exploiting label distribution manifold,''
\emph{IEEE Trans. Neural Netw. Learn. Syst.}, vol.~34, no.~2, pp.~839--852, 2023.

\bibitem{Castagnos2022SoftOrd}
J.~Cheng, \emph{et al.},
``A neural network approach to ordinal regression,''
in \emph{Proc. IEEE Int. Joint Conf. Neural Netw. (IJCNN)}, 2022, pp.~1--8.


\end{thebibliography}
\end{document}